\documentclass[letterpaper]{article} 
\usepackage{aaai23}  
\usepackage{times}  
\usepackage{helvet}  
\usepackage{courier}  
\usepackage[hyphens]{url}  
\usepackage{graphicx} 
\usepackage{natbib}  
\usepackage{caption} 
\usepackage{bm}
\usepackage{algorithm}
\usepackage{algorithmic}
\usepackage{multirow}
\usepackage{booktabs}
\usepackage{latexsym}
\usepackage{caption}
\usepackage{subcaption}
\usepackage{amsmath}
\usepackage{algorithm}
\usepackage{algorithmic}
\usepackage{extarrows}
\usepackage{amssymb}
\usepackage{multicol}
\usepackage{multirow}
\usepackage{amsmath}
\usepackage{color}
\usepackage{mathrsfs}
\usepackage{subcaption}
\usepackage{tabularx}
\usepackage{xcolor}
\usepackage{epstopdf}
\usepackage{graphicx}

\definecolor{bert_second}{HTML}{C8E3C8}

\definecolor{our_second}{HTML}{F1CBF1}

\usepackage{newfloat}
\usepackage{listings}
\DeclareCaptionStyle{ruled}{labelfont=normalfont,labelsep=colon,strut=off} 
\floatstyle{ruled}
\newfloat{listing}{tb}{lst}{}
\floatname{listing}{Listing}
\title{ProKD: An Unsupervised Prototypical Knowledge Distillation Network for Zero-Resource Cross-Lingual Named Entity Recognition}
\author{
    Ling Ge\textsuperscript{1},
    Chunming Hu\textsuperscript{1,2,}\thanks{Corresponding author},
    Guanghui Ma\textsuperscript{1},
    Hong Zhang\textsuperscript{3},
    Jihong Liu\textsuperscript{4}
}
\affiliations{
    \textsuperscript{\rm 1} School of Computer Science and Engineering, Beihang University, Beijing, China \\
    \textsuperscript{\rm 2} College of Software, Beihang University, Beijing, China \\
    \textsuperscript{\rm 3}National Computer Network Emergency Response Technical Team / Coordination Center of China, Beijing, China\\
    \textsuperscript{\rm 4}School of Mechanical Engineering and Automation, Beihang University, Beijing, China\\
    \{geling, hucm, maguanghui, ryukeiko\}@buaa.edu.cn, zhangh@isc.org.cn
}

\usepackage{amsmath}

\begin{document}

\maketitle

\begin{abstract}




For named entity recognition (NER) in zero-resource languages, utilizing knowledge distillation methods to transfer language-independent knowledge from the rich-resource source languages to zero-resource languages is an effective means. Typically, these approaches adopt a teacher-student architecture, where the teacher network is trained in the source language, and the student network seeks to learn knowledge from the teacher network and is expected to perform well in the target language. Despite the impressive performance achieved by these methods, we argue that they have two limitations. Firstly, the teacher network fails to effectively learn language-independent knowledge shared across languages due to the differences in the feature distribution between the source and target languages. Secondly,  the student network acquires all of its knowledge from the teacher network and ignores the learning of target language-specific knowledge.
Undesirably, these limitations would hinder the model's performance in the target language. This paper proposes an unsupervised prototype knowledge distillation network (ProKD) to address these issues. Specifically, ProKD presents a contrastive learning-based prototype alignment method to achieve class feature alignment by adjusting the distance among prototypes in the source and target languages, boosting the teacher network's capacity to acquire language-independent knowledge. In addition, ProKD introduces a prototypical self-training method to learn the intrinsic structure of the language by retraining the student network on the target data using samples' distance information from prototypes, thereby enhancing the student network's ability to acquire language-specific knowledge. Extensive experiments on three benchmark cross-lingual NER datasets demonstrate the effectiveness of our approach.

\end{abstract}
\section{Introduction}

Named Entity Recognition (NER) is a fundamental sub-task of information extraction that aims to locate and classify  text spans into  predefined entity classes such as locations, organizations, etc \cite{DBLP:conf/acl/MaJWZL22}. It is often employed as an essential component for  tasks such as question answering \cite{DBLP:conf/acl/CaoS0LY0L0X22} and coreference resolution\cite{DBLP:conf/acl/MaBDAMAR22}.  Despite the impressive performance recently achieved by deep learning-based NER methods, these supervised methods are limited to a few languages with rich entity labels, such as English, due to the reasonably large amount of human-annotated  training data required. In contrast, the total number of languages currently in use worldwide is about $7,000$ \footnote{https://www.ethnologue.com/guides/how-many-languages}, the majority of which contain limited or no labeled data, constraining the application of existing methods to these languages \cite{DBLP:conf/aaai/WuLWCKHL20, DBLP:conf/ijcai/WuLKHL20}. 
Hence, cross-lingual transfer learning is gaining increasing attention from researchers, which can leverage knowledge from high-resource (source) languages(e.g., English) with abundant entity labels to overcome the data scarcity problem of the  low- (zero-) resource (target) languages \cite{DBLP:conf/acl/LiuDBJSM20}. In particular, this paper focuses on the zero-resource scenario, where there is no labeled data in the target language.


To improve the performance of zero-resource cross-lingual NER, researchers have conducted intensive research and proposed various approaches\cite{DBLP:conf/emnlp/JainPL19, DBLP:conf/aaai/WuLWCKHL20, DBLP:conf/emnlp/PfeifferVGR20}. Among these, the knowledge distillation-based approaches \cite{DBLP:conf/acl/ChenJW0G20,DBLP:conf/ijcai/WuLKHL20, DBLP:conf/acl/WuLKLH20} have recently shown encouraging results.These approaches typically train a teacher NER network using source language data and then leverage the soft pseudo-labels produced by the teacher network for the target language data to train the student NER network. In this way, the student network is expected to learn the language-independent knowledge from the teacher network and perform well on unlabeled target data \cite{DBLP:journals/corr/HintonVD15}.

While significant progress has been achieved by knowledge distillation-based approaches for cross-lingual NER, we argue that these approaches still have two limitations.
First, knowledge distillation relies heavily on the shared language-independent knowledge acquired by the teacher network across languages. As is  known,  there are differences in the feature distribution between the source and target languages, existing techniques employ only the source language for teacher network training. As a result, the teacher network tends to learn source-language-specific knowledge and cannot effectively grasp shared language-independent knowledge. 
Second, under the knowledge distillation learning mechanism, the student network aims to match the pseudo-soft labels generated by the teacher network for the target language. Consequently, the student network acquires all of its knowledge from the teacher network and ignores the acquisition of target language-specific knowledge. Undesirably, these two constraints would hinder the model’s performance in the target language.

In this paper, we propose an unsupervised {\bf Pro}totypical {\bf K}nowledge {\bf D}istillation network ({\bf ProKD}), which employs contrastive learning-based prototype alignment and prototypical self-training to address the two above limitations, respectively. 
Specifically, we rely on performing class-level alignment between the source and target languages in semantic space to enhance the teacher network's capacity for capturing language-independent knowledge. We argue that class-level alignment can bridge the gap in the feature distribution and force the teacher network better to learn the shared semantics of entity classes across languages \cite{van2021crosslingual,DBLP:journals/corr/abs-2203-10444}. To do this, we choose prototypes \cite{DBLP:conf/nips/SnellSZ17}, i.e., the class-wise feature centroids, rather than samples, for class-level alignment because prototypes are robust to outliers and friendly to class imbalance tasks \cite{DBLP:conf/ijcai/Qiu0LNLDT21,DBLP:conf/cvpr/Zhang0Z0WW21}. 
In order to pull the prototypes of the same class closer and push the prototypes of different classes far away across languages, we leverage classical contrastive learning \cite{DBLP:conf/icml/ChenK0H20} to adjust the distance among class prototypes. Thus  the class-level representation alignment between the source and target languages is achieved.

Furthermore, we present a prototypical self-training method to enhance the student network's ability to acquire the target language-specific knowledge. In particular, we establish pseudo-hard labels for unlabeled target samples based on their softmax-valued relative distances, i.e., prototype probability, to all prototypes and then retrain the network using these pseudo-labels. Since the prototypes accurately represent the clustering distribution underlying the data, the prototypical self-training enables the student network to learn the intrinsic structure of the target language \cite{DBLP:conf/cvpr/Zhang0Z0WW21}, thus revealing language-specific knowledge, such as the token's label preference. In addition, while calculating the pseudo-hard labels, the class distribution probabilities generated by the teacher network are incorporated into the prototype probabilities to improve the quality of the pseudo-hard labels and facilitate self-training.

Summarily, we make four contributions: (1) We propose a ProKD model for zero-resource cross-lingual NER task, which can improve the model's generalization to the target language . (2) We propose a contrastive learning-based prototype alignment method to  enhance the teacher network's ability to acquire language-independent knowledge. (3) We propose a prototypical self-training method to enhance the student network's ability to acquire  target language-specific knowledge.  (4) Experimental results on  six target languages validate the effectiveness of our approach.

\section{Related Works}
\subsection{Cross-lingual NER}

Current research on cross-lingual NER with zero resources falls into three main branches.
The translation-based methods rely on machine translation and label projection \cite{DBLP:conf/emnlp/XieYNSC18,DBLP:conf/emnlp/JainPL19} to construct pseudo-training data for the target language, all of which involve high human costs and introduce label noise. 
The direct transfer-based methods resort to training a NER model with the source language and directly transferring it to the target language \cite{DBLP:conf/emnlp/WuD19, DBLP:conf/aaai/WuLWCKHL20, DBLP:conf/emnlp/PfeifferVGR20}. These approaches fail to exploit information from the unlabelled target language, resulting in non-optimal cross-lingual performance.
The knowledge distillation-based methods encourage the student network to learn language-independent knowledge from  the teacher network. Specifically, { \citet{DBLP:conf/acl/WuLKLH20}} distills knowledge directly from multi-source languages.
{ AdvPicker\cite{DBLP:conf/acl/ChenJW0G20}} leverages  adversarial learning to select  target data to alleviate the overfitting of the model to  source data. 
We argue that  the above approaches fail to effectively learn shared language-independent knowledge and ignore the acquisition of target language-specific knowledge.
\subsection{Knowledge Distillation}

Knowledge distillation enables knowledge transfer from the teacher network to the student network \cite{DBLP:journals/corr/HintonVD15}, where  the student network is optimized by fitting the soft labels generated by the trained teacher network. Since the soft targets have a high entropy value, they provide more information per training case than the hard targets \cite{DBLP:journals/corr/HintonVD15}, the student network can learn from the teacher network and perform well on unlabeled data. Knowledge distillation achieves significant results in various tasks such as model compression\cite{DBLP:conf/acl/LiuTF022}, image classification \cite{DBLP:journals/corr/HintonVD15}, dialogue generation \cite{DBLP:journals/corr/abs-1908-07137}, machine translation \cite{DBLP:conf/aaai/WengYHCL20}, etc. 
In this paper, we choose knowledge distillation as the basic framework of our proposed approach for zero-resource cross-lingual NER.

\section{Methodology}


 The NER task is modeled   as a sequence labeling problem in this paper, i.e., given a sentence $ {X}=\{x_0, \cdots, x_i,\cdots,x_L \}$, the NER model is expected to produce a label sequence $ {Y}=\{y_0, \cdots, y_i,\cdots,y_L \}$, where $y_i$ denotes the entity class corresponding to token $x_i$. Following previous works' setting\cite {DBLP:conf/acl/WuLKLH20,DBLP:conf/ijcai/WuLKHL20}, given a labeled dataset source language dataset, $ \{( { X}_m^s, { Y}_m^s)\}_{m=1}^{n_s} \sim \mathcal{D}_s$, and an unlabeled target language dataset, $ \{({ X}_m^t)\}_{m=1}^{n_t} \sim \mathcal{D}_t$,
the zero-resource cross-lingual NER aims to train a model with the above two datasets and expects the model to obtain good performance on target language data.




\subsection{Overall Architecture}

\begin{figure}[t!]
    \centering
    \includegraphics[width=0.48\textwidth]{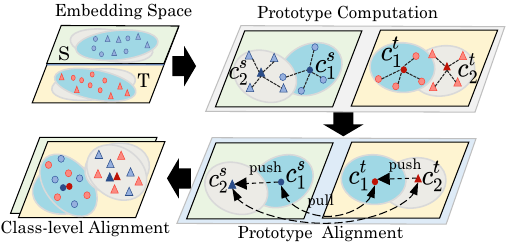}
\caption {We achieve class-level feature alignment  on the teacher network with a contrastive learning-based prototype alignment approach. The "S" and "T" denote the source and target languages, respectively.}
    \label{fig:teacher_network}
\end{figure}

\begin{figure}[t]
    \centering
    \includegraphics[width=0.47\textwidth]{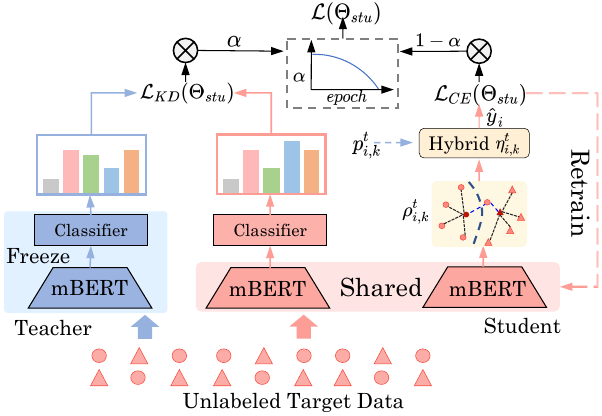}
    \caption{The student network can benefit from two aspects: the knowledge distillation and the self-training.}
    \label{fig:student_network}
\end{figure}

In this section, we describe the proposed approach, ProKD, for cross-lingual NER with zero resource, whose  architecture is shown in Fig \ref{fig:teacher_network} and Fig \ref{fig:student_network}.
The core of ProKD is a knowledge distillation framework that includes a teacher network and a student network. In more detail, the teacher network employs a prototype class alignment method based on contrastive learning, which enhances its ability to acquire language-independent knowledge. The student network utilizes a prototypical self-training approach combined with the class distribution probability of the teacher network, which enhances its ability to learn language-specific knowledge.

\subsection{Zero-resource Cross-Lingual NER via Knowledge Distillation}

The knowledge distillation-based methods for zero-resource cross-lingual NER typically follow a two-stage training pipeline.
First, the teacher network is trained with labeled source data, and then language-independent knowledge is distilled to the student network. 




Given a sequence $ {X_m^s}=\{x_0^s, \cdots, x_i^s,\cdots,x_L^s \}$ from source language data, the encoder $f_\theta$ of teacher network can  map it into the hidden space and output the representations $ {H^s_m}=\{h_0^s, \cdots, h_i^s,\cdots,h_L^s \}$. Following the  previous works \cite{ DBLP:conf/acl/WuLKLH20,DBLP:conf/ijcai/WuLKHL20}, we adopt multilingual BERT(short for mBERT) \cite{DBLP:conf/naacl/Defunct19} as the feature encoder. Then we leverage a classifier with a softmax function to obtain the  output $p_i^s$ for each token $x_i^s$, and  the cross entropy loss  for the teacher network can be  formulated as:


 



\begin{equation}
\begin{split}
\mathcal L_{CE}(\Theta_{tea})
=-\frac{1}{n_s L}  
 \sum_{(x^s,y^s) \in D_s}  \sum_{i=0}^{L} y_i^s log(p_i^s )
\end{split}
\label{ce_tea}
\end{equation}
where $\Theta_{tea}$ is the parameters of the teacher network to be optimized,   $n_s$ is the number of the sentences  in dataset $D_s$, and $y_i^s$ represents the golden label of token $x_i^s$.

Benefiting from the shared feature space of pre-trained mBERT and task knowledge from the labeled source data, we can directly utilize the teacher network to infer the class probabilities $p_i^t$ of each token in a sequence ${X_m^t}$ from unlabeled dataset ${D}_t^x$.
Then the student network,  consisting of a feature encoder mBERT and a classifier with a softmax function, is trained using these class probabilities as "soft targets" on the unlabeled dataset.
To approximate the probabilities $p_i^t$ , the training objective for the student network can be  formulated as:

\begin{equation} 
\begin{split}
\mathcal{L}_{KD}(\Theta_{stu}) =\frac{1}{L}\sum_{x \in D_t} \sum_{i=1}^L  ( p_i^t - q_i^t)^2
\end{split}
\label{loss_con}
\end{equation}
where  $ p_i^t$ and $ q_i^t$ denote the probability distribution produced by the teacher and the student network for $x_i^t$, respectively. And here, following previous works \cite{DBLP:conf/wsdm/YangSGLJ20, DBLP:conf/acl/WuLKLH20}, we use the MSE loss to measure the prediction discrepancy of the two networks.








\subsection{Prototypical  Class-wise Alignment}



Here, we present our method, prototypical class-wise alignment, to boost the teacher network’s capacity to acquire language-independent knowledge.


 




Due to the absence of  annotations on target language data, the class-wise alignment between the source and target languages is not trivial. To address this, as shown in Fig \ref{fig:teacher_network}, we first calculate target class prototypes by class distribution probabilities produced by the teacher network in target data, and then leverage the prototype alignment between the two above  languages to achieve class-wise alignment. We use prototype alignment rather than sample alignment since the prototype is robust to outliers \cite{DBLP:conf/cvpr/Zhang0Z0WW21}, and it can alleviate the negative impact of the noise \cite{DBLP:conf/icml/XieZCC18} introduced by the teacher network for the target data. Additionally, the prototype treats all classes equally \cite{DBLP:conf/cvpr/Zhang0Z0WW21}, which is crucial for the NER task, as non-entity type samples constitute the bulk of the overall samples.





To be specific, for the source language, we first obtain the token representation $h_i^s$ of  each token $x_i^s$ using mBERT, and then, with the help of the golden labels,  we directly compute the average representation of token samples with  same label and treat it as the class prototype:

\begin{equation}
\begin{split}
C_k^s=\frac{1}{n_k^s} \sum_{(X^s,Y^s) \in D_s}\sum_{i=0}^L \mathbb{I}[y_i^s = k] h_i^s
\end{split}
\label{loss_con}
\end{equation}
where $k$ denotes an entity class label, $\mathbb{I}$ is an indicator function, and $n_k^s$ represents the number of samples belonging to class $k$ in the source language.


For the target language, we utilize the same method to obtain the representation $h_i^t$ of  each target token $x_i^t$. Since the target data is unlabeled, to alleviate the uncertainty of class prototype  computation, we use the output of the teacher classifier to estimate the  probabilities for the current token belonging to each class. Regarding these probabilities as weight, we  aggregate representations of all target tokens to derive the target class prototype, which can be expressed as:



\begin{equation}
\begin{split}
 C_k^t=\frac{\sum_{X \in D_t} \sum_{i=0}^L p_{i,k}^t * h_i^t}{\sum_{X \in D_t} \sum_{i=0}^L  p_{i,k}} 
\end{split}
\label{proto_com}
\end{equation}
where $p_{i,k}^t$  represents the probability that the token $x_i$ belongs to the class $k$.

The class prototype calculation involves  all the samples, leading to high computing costs. To reduce the computation complexity while ensuring the stability of updates,
we use the moving average method \cite{DBLP:conf/icml/XieZCC18} to update the source and target prototypes :
\begin{equation}
\begin{split}
\mathcal C_{k,cur}^{s(t)}= \lambda * C_{k,cur}^{s(t)}+(1-\lambda )* C_{k,cur-1}^{s(t)}
\end{split}
\label{proto_moving}
\end{equation}
where $\lambda \in (0,1)$ is the moving average coefficient, $cur$ denotes the current moment and $cur-1$ indicates the previous moment. In  practical implementation, the source prototypes are updated once per epoch, while the target prototypes are updated once per batch.



After obtaining  all class prototypes, we leverage classical contrastive learning to adjust the distance among prototypes in the feature space for class-wise alignment.
For prototypes from source and target data with the same class, we regard one as an anchor (e.g., $ C_i^s$ ) and the other as the positive sample of the anchor (e.g., $ C_i^t$ ), while the rest of the prototypes are considered as negative samples (marked as $ C_{i,neg}^s$ ). Then the class alignment loss is presented as:




\begin{equation}
\begin{split} 
& { \mathcal L_{CA}(\Theta_{tea})} = -\log \sum_{i=1}^{num}
 \\& 
{ \frac{ \exp ({z_i^s  \cdot z_i^t} /  \tau_1 ) } {   \sum_{neg  } \exp ( {z_i^s  \cdot  z_{i,neg}^s} /  \tau_1 ) + \sum_{neg  } \exp ( {z_i^t  \cdot  z_{i,neg}^t} /  \tau_1 )} }
\end{split}
\end{equation}
where $z_i^s$ , $z_i^t$, $z_{i,neg}^s$ and $z_{i,neg}^t$ are $l_2$ regularization of $C_i^s$ , $C_i^t$, $C_{i,neg}^s$, $C_{i,neg}^t$, respectively, the  $C_{i,neg}^t$ denotes the negative samples of  $C_i^t$,  $\tau_1$ is a temperature parameter, and  $num$ is the number of entity classes . 



In this way, we can pull in source and target prototypes of the same class and push away source and target prototypes of different classes. 
Finally, we obtain the total loss $ \mathcal L(\Theta_{tea})$ for the teacher network, consisting of the cross-entropy loss and the class alignment loss:


\begin{equation}
\begin{split}
 \mathcal L(\Theta_{tea})  =  \mathcal L_{CE}(\Theta_{tea}) +  \mathcal L_{CA}(\Theta_{tea})
\end{split}
\label{loss_tea}
\end{equation}





\subsection{Prototypical Self-trainning}



Here, we present our approach prototypical self-training with the unlabeled target language data, 
to boost the student network’s ability to learn
language-specific knowledge.

Specifically, we rely on prototype learning to iteratively generate hard pseudo  labels for unlabelled target language samples  and leverage these hard labels to conduct self-training on the target data. This is because the prototypes can perceive the underlying clustering distribution of the data, fundamentally reflecting the internal structure of the data and the intrinsic differences across the data \cite{DBLP:conf/cvpr/Zhang0Z0WW21}, which facilitates the learning of language-specific knowledge, such as the label preference of a token.



To acquire the target class prototypes, we first obtain the hidden representations and prediction probabilities  through the student network, respectively, and then leverage the exact prototype computation and updating equation  (Equation \ref{proto_com} and Equation \ref{proto_moving}) as the teacher network to obtain the class prototypes $ C^t$.
Afterwards,  a class probability distribution $\rho_i$ based on prototypes is calculated by leveraging the  sample’s feature distance w.r.t the class prototypes:

\begin{equation}
\begin{split}
 \rho_{i,k}^t= \frac{exp(-\Vert h_i^t - C_k^t \Vert / \tau_2)}{\sum_{k'} exp(-\Vert h_i^t -  C_{k'}^t \Vert / \tau_2)}
\end{split}
\label{proto_distance}
\end{equation}
where $\tau_2$  is the softmax temperature, and $\rho_{i,k}^t$ represents the softmax probability of sample $x_i$ belonging to the $k$th class. As observed, if  a feature representation $ h_i^t $ is  far from the prototype $ C_k^t$, the  probability of this feature for class $k$ would  be very low.
We convert $\rho_i^t$ into a hard pseudo-label $\hat y_i^t$ based on the following formula:

\begin{equation}
\begin{split}
 \hat y_i^t= \xi (\rho_{i}^t)
\end{split}
\label{argmax_selftrain}
\end{equation}
where $\xi$ denotes the conversion function.


Intuitively, we can use these pseudo-hard labels for self-training. However, one natural question then arises. Prototypical self-training is essentially cluster-based representation learning and will inevitably introduce incorrect label in pseudo labeling. For instance, when a sample is far from the prototype to which it belongs, the student network may mislabel this sample\cite{DBLP:conf/nips/SnellSZ17}.
To alleviate this issue, we fuse the above prototypical probability $\rho_{i,k}^t$ with the teacher's output probability $p_{i,k}^t$, to produce a  hybrid soft pseudo-label $\eta_{i,k}^t$:

\begin{equation}
\begin{split}
 \eta_{i,k}^t=  \gamma *  \rho_{i,k}^t + (1-\gamma) * p_{i,k}^t
\end{split}
\label{proto_distance}
\end{equation}
where $\gamma$ is a fuse factor.

Since the trained teacher has the  general semantic knowledge of classes, the $p_{i,k}^t$ can be regarded as a priori knowledge, to improve the quality of pseudo labeling, which shows appealing advantages in previous works  \cite{DBLP:conf/iclr/0001XH21, DBLP:conf/cvpr/Zhang0Z0WW21}.


Note that, the teacher network's output $p_{i,k}^t$  remains fixed as training proceeds. The reason we choose  $p_{i,k}^t$ instead of the updating probability $q_{i,k}^t$  of the student, is to avoid the degenerate solution, resulting from the simultaneous update of features and labels throughout the self-training.
Subsequently, we use the hybrid  $\eta_i^t$  instead of  $\rho_i^t$ to produce pseudo hard label.
To this end, the student can be trained by the traditional self-training loss \cite{DBLP:conf/eccv/ZouYKW18}:
\begin{equation}
\begin{split}
 \mathcal L_{CE}(\Theta_{stu})= -\sum_{x\in D_t^x} \sum_{i=1}^L \xi(\eta_{i}^t) log(q_{i}^t)
\end{split}
\label{self_train_loss}
\end{equation}




where $ q_i^t$ denotes the probability distribution produced via the classifier of the student network for $x_i^t$.

Based on the above, the student network can benefit from two aspects (Fig \ref{fig:student_network}) : knowledge distillation and self-training.
A very straightforward issue is that the student network may not be competent at an early stage to undertake effective self-training .
To guarantee that the student network can learn the shared class semantic  for self-training at the early stage , we follow a cumulative learning strategy \cite{DBLP:conf/cvpr/ZhouCWC20} to gradually shift the model's learning focus from knowledge distillation to self-training using the control parameter $\alpha$ : 


\begin{equation}
\begin{split}
 \alpha = 1 - (\frac{e}{E_{max}})^2
\end{split}
\label{self_train_loss}
\end{equation}
 where $E_{max}$ is the number of total training epochs, and $e$ is  the current epoch. The  $ \alpha $ automatically decreases from 1 to 0 with increasing epoch.

Finally, the  loss $\mathcal L(\Theta_{stu})$ for the student network can be expressed as:

%
\begin{equation}
\begin{split}
 \mathcal L(\Theta_{stu})=(1-\alpha )  \mathcal L_{CE}(\Theta_{stu}) + \alpha \mathcal L_{KD}(\Theta_{stu})
\end{split}
\label{loss_stu}
\end{equation}




\section{Experiments and Analysis}

\subsection{Datasets}

We adopt three widely-used benchmark datasets for experiments: {\bf CoNLL-2002} (Spanish and Dutch) \cite{DBLP:conf/conll/Sang02}, {\bf CoNLL-2003}  (English and German) \cite{DBLP:conf/conll/SangM03}, and {\bf Wikiann} (English, Arabic, Hindi and Chinese) \cite{DBLP:conf/acl/PanZMNKJ17}.
Each language-specific dataset has the standard training, development, and evaluation sets.The statistics for all datasets are shown in Table \ref{tab:statistics}. 

Following previous works \cite{DBLP:conf/acl/WuLKLH20,DBLP:conf/aaai/WuLWCKHL20}, we apply word-piece \cite{DBLP:journals/corr/WuSCLNMKCGMKSJL16} to tokenize the sentences into sub-words, which then be marked by the BIO scheme. The data are annotated with four different entity types: PER (Persons), LOC (Locations), ORG (Organizations), and MISC (Miscellaneous). For all experiments, English is regarded as the source language and  others as the target language respectively. Note that, CoNLL-2002/2003 share a common English dataset as source data. Moreover, we train the model on the source language training set, validate the model  on the source language development set, and evaluate the learned model on the target language test set to simulate the zero-resource cross-language NER scenario.

\begin{table}[]
\renewcommand\arraystretch{1.1}
\setlength\tabcolsep{3.0pt}
    \centering
\begin{tabular}{cccccc}
\toprule[1.0pt]
\textbf{Datasets}                               & \textbf{Language}                                                       & \textbf{Type} & \textbf{Train} & \textbf{Dev} & \textbf{Test} \\ \hline
\multicolumn{1}{c|}{\multirow{4}{*}{Conll2003}} & \multirow{2}{*}{\begin{tabular}[c]{@{}c@{}}English\\ (en)\end{tabular}} & Sentence      & 14,987          & 3,466         & 3,684          \\
\multicolumn{1}{c|}{}                           &                                                                         & Entity        & 23,499          & 5,942         & 5,648          \\ \cline{2-6} 
\multicolumn{1}{c|}{}                           & \multirow{2}{*}{\begin{tabular}[c]{@{}c@{}}German\\ (de)\end{tabular}}  & Sentence      & 12,705          & 3,068         & 3,160          \\
\multicolumn{1}{c|}{}                           &                                                                         & Entity        & 11,851          & 4,833         & 3,673          \\ \hline
\multicolumn{1}{c|}{\multirow{4}{*}{Conll2002}} & \multirow{2}{*}{\begin{tabular}[c]{@{}c@{}}Spanish\\ (es)\end{tabular}} & Sentence      & 8,323           & 1,915         & 1,517          \\
\multicolumn{1}{c|}{}                           &                                                                         & Entity        & 18,798          & 4,351         & 3,558          \\ \cline{2-6} 
\multicolumn{1}{c|}{}                           & \multirow{2}{*}{\begin{tabular}[c]{@{}c@{}}Dutch\\ (nl)\end{tabular}}   & Sentence      & 15,806          & 2,895         & 5,195          \\
\multicolumn{1}{c|}{}                           &                                                                         & Entity        & 13,344          & 2,616         & 3,941          \\ \hline
\multicolumn{1}{c|}{\multirow{8}{*}{Wikiann}}   & \multirow{2}{*}{\begin{tabular}[c]{@{}c@{}}English\\ (en)\end{tabular}} & Sentence      & 20,000          & 10,000        & 10,000         \\
\multicolumn{1}{c|}{}                           &                                                                         & Entity        & 27,931          & 14,146        & 13,958         \\ \cline{2-6} 
\multicolumn{1}{c|}{}                           & \multirow{2}{*}{\begin{tabular}[c]{@{}c@{}}Arabic\\ (ar)\end{tabular}}  & Sentence      & 20,000          & 10,000        & 10,000         \\
\multicolumn{1}{c|}{}                           &                                                                         & Entity        & 22,500          & 11,266        & 11,259         \\ \cline{2-6} 
\multicolumn{1}{c|}{}                           & \multirow{2}{*}{\begin{tabular}[c]{@{}c@{}}Hindi\\ (hi)\end{tabular}}   & Sentence      & 5,000           & 1,000         & 1,000          \\
\multicolumn{1}{c|}{}                           &                                                                         & Entity        & 6,124           & 1,226         & 1,228          \\ \cline{2-6} 
\multicolumn{1}{c|}{}                           & \multirow{2}{*}{\begin{tabular}[c]{@{}c@{}}Chinese\\ (zh)\end{tabular}} & Sentence      & 20,000          & 10,000        & 10,000         \\
\multicolumn{1}{c|}{}                           &                                                                         & Entity        & 25,031          & 12,493        & 12,532         \\ \toprule[1.0pt]
\end{tabular}

    \caption{Statistics of the datasets.}
    \label{tab:statistics}
\end{table}

\begin{table}[t]
\setlength\tabcolsep{3.2pt}
    \centering
\begin{tabular}{ccccc}
\toprule[1.0pt]
\textbf{Method} & \textbf{es} & \textbf{nl} & \textbf{de} & \textbf{Avg} \\ \hline
\citet{DBLP:conf/acl/NiDF17}  & 65.10       & 65.40       & 58.50       & 63.00            \\
\citet{DBLP:conf/emnlp/MayhewTR17} & 65.95       & 66.50       & 59.11       & 63.85            \\
\citet{DBLP:conf/emnlp/XieYNSC18}  & 72.37       & 71.25       & 57.76       & 67.13            \\
\citet{DBLP:conf/emnlp/WuD19}  & 74.50       & 79.50       & 71.10       & 75.03            \\
\citet{DBLP:journals/corr/abs-1912-01389}  & 75.67       & 80.38       & 71.42       & 75.82            \\
\citet{DBLP:conf/aaai/WuLWCKHL20}  & 76.75       & 80.44       & 73.16       & 76.78            \\
\citet{DBLP:conf/acl/WuLKLH20} & 76.94       & 80.99       & 73.22       & 77.02            \\
UniTrans \cite{DBLP:conf/ijcai/WuLKHL20}  & 77.30       & 81.20       & 73.61       &77.37            \\
RIKD\cite{DBLP:conf/kdd/LiangGPSZZJ21} & 77.84       & 82.46       & 75.48       & 78.59            \\
AdvPicker\cite{DBLP:conf/acl/ChenJW0G20}& 79.00       & \textbf{82.90}       & 75.01       & 78.97            \\
\textbf{ProKD (Ours)}            & \textbf{79.53}         &82.62           & \textbf{78.90}         &\textbf{80.35}                \\
\toprule[1.0pt]
\end{tabular}
    \caption{Result comparison on Conll2002\&2003.}
\label{tab:conll_result}
\end{table}

\begin{table}[]
\setlength\tabcolsep{4.0pt}
    \centering
\begin{tabular}{ccccc}
\toprule[1.0pt]
\textbf{Method} & \textbf{ar} & \textbf{hi} & \textbf{zh} & \textbf{Avg} \\ \hline
     \citet{DBLP:conf/emnlp/WuD20}            & 42.30       & 67.60       & -       & -            \\
      \citet{DBLP:conf/acl/WuLKLH20}           & 43.12       & 69.54       & 48.12       & 53.59            \\ 
      RIKD \cite{DBLP:conf/kdd/LiangGPSZZJ21} 
                & 45.96       & 70.28       & 50.40       & 55.55            \\
\textbf {ProKD (Ours)}              &  \textbf{50.91}           &  \textbf{70.72}           &  \textbf{51.80}           &  \textbf{57.81}                \\ \toprule[1.0pt]
\end{tabular}
    \caption{Result comparison  on Wikiann. }
    \label{tab:Wikiann_result}
\end{table}

\begin{figure*}
     \centering
          \begin{subfigure}[b]{0.90\textwidth}
         \centering
         \includegraphics[width=\textwidth]{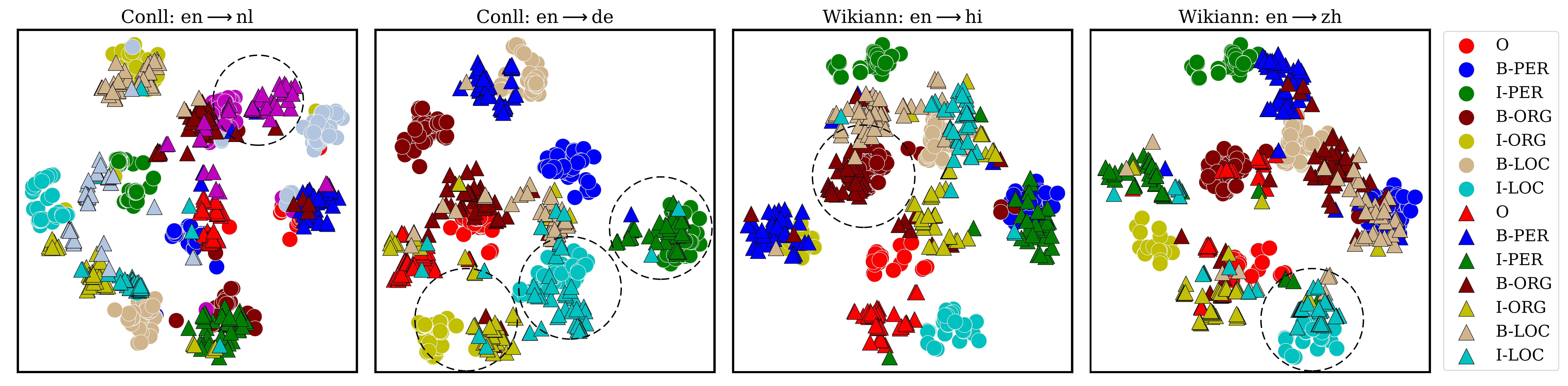}
         \caption{ProKD$_{w/o \; \rm{CA}}$}
         \label{subfig:mBERT}
     \end{subfigure}
          \begin{subfigure}[b]{0.90\textwidth}
         \centering
         \includegraphics[width=\textwidth]{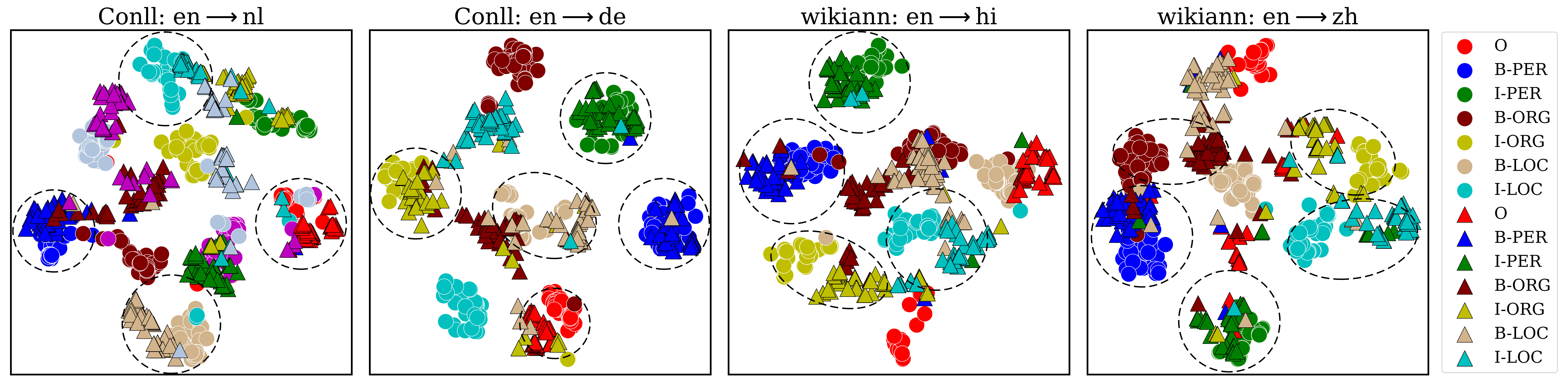}
         \caption{ProKD}
         \label{subfig:Ours}
     \end{subfigure}
        \caption{ Our ProKD shows superiority over ProKD$_{w/o \; \rm{CA}}$ with more classes aligned correctly. The circles ($\bullet$) and triangles ($\blacktriangle$) indicate sample representations of the source and target languages, respectively. Dashed circles indicate that samples from different languages belonging to the same class are correctly aligned.}
        \label{fig:Visualing}
\end{figure*}

\subsection{Implementation Details}
We adopt the pre-trained mBERT\cite{DBLP:conf/acl/PiresSG19}  as the feature extractor. Following previous works \cite{DBLP:conf/acl/WuLKLH20,DBLP:conf/aaai/WuLWCKHL20}, we use the token-level F1 score as the evaluation metric.
For all experiments, we use  the Adam optimizer \cite{DBLP:journals/corr/KingmaB14} with learning rate =  5e-5 for teacher network and 1e-5 for student network, batch size = 128, maximum sequence
length = 128, and the dropout = 0.5 empirically. We utilize the grid search technology to obtain the optimal super-parameters, including the  moving average coefficient $\lambda$ selected from $\{0.001,0.005, 0.0001, 0.0005\}$, the contrastive learning temperature $\tau_1$ selected  from $0.5$ to $0.9$, the softmax temperature $\tau_2$ selected  from  $0.5$ to $0.9$, and the fuse factor $\gamma$ selected  from  $0.7$ to $0.9$.

Following the previous work \cite{DBLP:conf/emnlp/WuD20},  we only consider the first sub-word tokenized by word-piece  in our loss function  and freeze the parameters of the embedding layer and the bottom three layers of the
mBERT model.
Additionally, our approach is implemented using PyTorch, and all calculations are done on NVIDIA Tesla V100 GPU.

\begin{table}[t]
    \centering
    \setlength\tabcolsep{4.0pt}
\begin{tabular}{lcccc}
\toprule[1.0pt]
\multicolumn{5}{c}{Conll 2022 \& 2003} \\ \hline
   Method  & es    & nl    & de    & average \\ \hline
ProKD &  \textbf{79.53}         &\textbf{82.62}           & \textbf{78.90}         &\textbf{80.35}      \\
{ProKD$_{w/o \; \rm{CA}}$ } & 77.46 & 80.34 &77.36 & 78.25 ($\downarrow$2.10 )   \\
{ProKD$_{w/o  \; \rm{ST}}$ } & 77.85& 80.69 & 77.85 & 78.80 ($\downarrow$1.55 )   \\
{ProKD$_{w/o  \; \rm{PK}}$ }  & 79.35 & 82.00 &78.56& 79.97 ($\downarrow$0.38 )   \\
{ProKD$_{w/o  \; \rm{CL}}$ } & 79.42 & 82.20 & 78.63 & 80.08 ($\downarrow$0.27 )  \\ 
\toprule[0.7pt]
\multicolumn{5}{c}{Wikiann}            \\
\hline
Method     & ar    & hi    & zh    & average \\ \hline
ProKD & \textbf{50.91}           &  \textbf{70.72}           &  \textbf{51.80}           &  \textbf{57.81}   \\
{ProKD$_{w/o \; \rm{CA}}$ } & 48.88           &69.02         & 49.57          & 55.82  ($\downarrow$1.99 ) \\
{ProKD$_{w/o  \; \rm{ST}}$ } & 49.61 &69.57 & 50.12 & 56.43 ($\downarrow$1.38 )  \\
{ProKD$_{w/o  \; \rm{PK}}$ }  & 50.45 & 70.41 &51.33 & 57.40 ($\downarrow$0.41 )  \\
{ProKD$_{w/o  \; \rm{CL}}$ } &50.73 & 70.66 & 51.56 & 57.65($\downarrow$0.16 )   \\ 
\toprule[1.0pt]
\end{tabular}
    \caption{Ablation study on different factors.}
\label{tab:abala}
\end{table}

\begin{table*}[t]
\renewcommand\arraystretch{1.5}
\setlength\tabcolsep{3.5pt}
    \centering
\begin{tabular}{c|l|c|c}
\hline
\multicolumn{1}{l|}{\textbf{}}                                         & \multicolumn{1}{c|}{\textbf{test text}}                                           & \textbf{source}                                                                                    & \textbf{target}                                                                                    \\ \hline
\multirow{2}{*}{\begin{tabular}[c]{@{}c@{}}\#1\\ Spanish\end{tabular}} & {ProKD$_{w/o  \; \rm{ST}}$ }: Para UGT de
\colorbox{our_second}{\color[rgb]{0,0,0}Madrid {[}I-ORG{]}}, la decisión de la UE ...          & \multirow{2}{*}{\begin{tabular}[c]{@{}c@{}}66.67\% {[}I-ORG{]}\\ 16.67\% {[}B-LOC{]}\end{tabular}} & \multirow{2}{*}{\begin{tabular}[c]{@{}c@{}}32.16\% {[}I-ORG{]}\\ 59.73\% {[}B-LOC{]}\end{tabular}} \\
                                                                       & {ProKD}: Para UGT de \colorbox{bert_second}{\color[rgb]{0,0,0}Madrid {[}B-LOC{]}}, la decisión de la UE ...          &                                                                                                    &    \\ \hline

\multirow{2}{*}{\begin{tabular}[c]{@{}c@{}}\#2\\ German\end{tabular}}  & {ProKD$_{w/o  \; \rm{ST}}$ }: Sozialdezernent 
\colorbox{our_second}{\color[rgb]{0,0,0}Martin {[}I-PER{]}} Berg war "vor Ort" in... & \multirow{2}{*}{\begin{tabular}[c]{@{}c@{}}53.19\% {[}B-PER{]}\\ 46.80\% {[}I-PER{]}\end{tabular}} & \multirow{2}{*}{\begin{tabular}[c]{@{}c@{}}84.38\% {[}B-PER{]}\\ 9.38\% {[}I-PER{]}\end{tabular}}  \\
                                                                       &{ProKD }: Sozialdezernent
                                                                       \colorbox{bert_second}{\color[rgb]{0,0,0}Martin {[}B-PER{]}} Berg war "vor Ort" in ... &                                                                                                    &                                                                                                                                                                    \\ \hline
\multirow{2}{*}{\begin{tabular}[c]{@{}c@{}}\#3\\ Dutchc\end{tabular}}  & {ProKD$_{w/o  \; \rm{ST}}$ }: we dit najaar zullen hebben met \colorbox{our_second}{\color[rgb]{0,0,0}Vandenbroucke{[}I-PER{]}} ...         & \multirow{2}{*}{\begin{tabular}[c]{@{}c@{}}100.00\% {[}I-PER{]}\\ 0.00\% {[}B-PER{]}\end{tabular}} & \multirow{2}{*}{\begin{tabular}[c]{@{}c@{}}33.33\% {[}I-PER{]}\\ 66.67\% {[}B-PER{]}\end{tabular}} \\
                                                                       & ProKD:we dit najaar zullen hebben met \colorbox{bert_second}{\color[rgb]{0,0,0} Vandenbroucke {[}B-PER{]}}...        &                                                                                                    &                                                                                                    \\ \hline
\end{tabular}
\caption{Case Study. The ProKD can learn language-specific knowledge with self-training, which helps the model to rectify  \colorbox{our_second}{\color[rgb]{0,0,0}incorrect} predictions to  \colorbox{bert_second}{\color[rgb]{0,0,0}correct} ones.}
\label{table:case}
\end{table*}

\subsection{Performance Comparison}


We compare the proposed approach with several previous approaches, including three translation-based approaches: {\bf \citet{DBLP:conf/emnlp/MayhewTR17}},  {\bf \citet{DBLP:conf/acl/NiDF17}}and {\bf \citet{DBLP:conf/emnlp/XieYNSC18} }, three direct tranfer-based approaches: {\bf \citet{DBLP:conf/emnlp/WuD19}}, {\bf \citet{DBLP:journals/corr/abs-1912-01389}} and {\bf \citet{DBLP:conf/aaai/WuLWCKHL20}}, and four knowledge distillation-based approaches: {\bf \citet{DBLP:conf/acl/WuLKLH20}},  {\bf UniTrans \cite{DBLP:conf/ijcai/WuLKHL20}}, {\bf RIKD\cite{DBLP:conf/kdd/LiangGPSZZJ21}} and {\bf AdvPicker\cite{DBLP:conf/acl/ChenJW0G20}}.

The results are presented in Table \ref{tab:conll_result} and Table \ref{tab:Wikiann_result} ,where the baseline and SOTA experimental results are from their original papers. As observed, our method achieves the best results on most the datasets. 
For Conll2002/2003, compared with the two competitive knowledge distillation-based methods, RIKD and AdvPicer, our approach improves on the average F1  by 1.76\% and 1.38\%,  respectively. For Wikiann, our method outperforms the RIKD  by 2.26\% on average. 
Especially, for German(de) language, we obtain an F1 value of 78.9\%, which is 3.42\% higher than the best result of the RIKD.  And for Arabic(ar) language, our method achieves the best F1  value of 50.91\%, with an improvement of 4.95\% than RIKD. 
Analytically, RIKD and AdvPicer leverage adversarial learning and reinforcement learning to select target data for distillation, respectively, and the  selected data tends to be consistent with the source language in feature distribution. Consequently, the student network learning on this data fail to effectively acquire  the target language knowledge, resulting in insufficient generalization on the target language. Contrastly, our model uses the prototypical self-training to enhance the student network's ability to learn the target language, thus performing well on the target language.

\subsection{Ablation Study}
To investigate the contributions of different factors, we conduct ablation experiments with four variant models: (1) {\bf ProKD$_{w/o \; \rm{CA}}$ } removes the prototypical class-wise alignment from the teacher network. (2) {\bf ProKD$_{w/o  \; \rm{ST}}$ } wipes out the prototypical self-training from the student network. (3) {\bf ProKD$_{w/o  \; \rm{PK}}$ } does not use the prior knowledge from the teacher network  in self-training process. (4) {\bf ProKD$_{w/o  \; \rm{CL}}$ } cuts out the cumulative learning scheme and adopts the parameter $\alpha =0.5$ in  loss function (Equation \ref{loss_stu}) for student network.
As shown in Table \ref{tab:abala},  the average F1 value of ProKD$_{w/o \; \rm{CA}}$ decrease by 2.1\% compared to ProKD on Conll 2002 \& 2003. This indicates that class-level alignment effectively improves  the model's generalization, as class alignment forces the teacher network to learn language-independent  knowledge from source and target languages. The performance of the ProKD$_{w/o \; \rm{ST}}$ in  F1 score drops by 1.55\% compared to ProKD, which well validates the effectiveness of the self-training  to  acquire  the target language-specific knowledge.  For ProKD$_{w/o  \; \rm{PK}}$, the slight drop in F1 results compared to the ProKD suggests that incorporating prior knowledge of the teacher network can  enhance the quality of pseudo labels  during self-training. Also, ProKD$_{w/o  \; \rm{CL}}$ yields a slight drop in F1 values,  which proves that the knowledge distillation learning should be performed first and then the self-training. The above experimental phenomena can also be observed on the Wikiann dataset.




\subsection{Visualizing the Token Sample Representations}
To demonstrate that our ProKD can achieve class-level feature alignment, we randomly select 50 token samples for each class from the source and target languages and feed them to the teacher networks of  ProKD and ProKD$_{w/o \; \rm{CA}}$ to obtain token-level representations, respectively. Note that, the teacher network of ProKD$_{w/o \; \rm{CA}}$ degenerates to a vanilla mBERT when removing  the prototypical class-wise alignment. We then visualize these representations  using the T-SNE \cite{van2008visualizing}  and show the results for the four target languages in Figure \ref{fig:Visualing}. As shown,  the  feature  representations of source and target languages  from ProKD$_{w/o \; \rm{CA}}$ are distributed differently and inconsistent due to languages gap. Many target language examples of one class are incorrectly aligned to the  source language examples of a different class, thus causing confusion and hindering the model's performance. By contrast, our approach ProKD shows superiority over ProKD$_{w/o \; \rm{CA}}$ with more classes aligned correctly. 
For example, when performing a cross-lingual NER from English(en) to Chinese(zh), the  ProKD$_{w/o \; \rm{CA}}$ aligns source and target features for just one class, ILOC, while our model achieves feature alignment on five classes. We argue that a model aligning features across multiple classes can capture more shared class features across languages,  which is essential for generalizing the model to unknown target languages.

\subsection{Case Study}

In this part, we present a case study to show that our model can learn target language-specific knowledge through self-training. We compare the prediction results of the ProKD$_{w/o \; \rm{ST}}$  with our ProKD for the target language test data, as shown in Table \ref{table:case}. In example $1$, the ProKD$_{w/o \; \rm{ST}}$ model incorrectly predicts "Madrid" as "I-ORG" because 66.67\% of the "Madrid" tokens in  English dataset are annotated as "I-ORG". The teacher network trained with this English data would distill this label preference to the student network, resulting in the student network of ProKD$_{w/o \; \rm{ST}}$ tending to make incorrect predictions.
In contrast, our model captures the label preferences of the target language by a prototypical self-training learning mechanism. In the same example $1$,  59.73\% of "Madrid" tokens in target Spanish language are labeled as "I-ORG". Our model can produce accurate predictions due to its intimate familiarity with the target language-specific knowledge.  We observe the same phenomenon in examples $2$ and $3$.

\section{Conclusion}
This paper presents a knowledge distillation-based  network ProKD for zero-resource cross-lingual NER. The ProKD proposes a  contrastive learning-based prototype alignment approach to boost the teacher network's capacity to capture language-independent knowledge. In addition, the  ProKD introduces the prototypical self-training method to improve the student network's capacity to grasp language-specific of target knowledge. The experiments on six target languages illustrate the effectiveness of the proposed approach.

\section{Acknowledgments}
We sincerely thank the reviewers for their insightful comments and valuable suggestions. This work is funded by the National Key Research and Development Program of the Ministry of Science and Technology of China (No. 2021YFB1716201). Thanks for the computing infrastructure provided by Beijing Advanced Innovation Center for Big Data and Brain Computing.
\bibliography{aaai23}



\end{document}